# Toward Enabling a Reliable Quality Monitoring System for Additive Manufacturing Process using Deep Convolutional Neural Networks

Yaser Banadaki, Nariman Razaviarab, Hadi Fekrmandi, and Safura Sharifi

*Abstract*—Additive Manufacturing (AM) is a crucial component of the smart industry. In this paper, we propose an automated quality grading system for the AM process using a deep convolutional neural network (CNN) model. The CNN model is trained offline using the images of the internal and surface defects in the layer-by-layer deposition of materials and tested online by studying the performance of detecting and classifying the failure in AM process at different extruder speeds and temperatures. The model demonstrates the accuracy of 94% and specificity of 96%, as well as above 75% in three classifier measures of the F-score, the sensitivity, and precision for classifying the quality of the printing process in five grades in real-time. The proposed online model adds an automated, consistent, and non-contact quality control signal to the AM process that eliminates the manual inspection of parts after they are entirely built. The quality monitoring signal can also be used by the machine to suggest remedial actions by adjusting the parameters in real-time. The proposed quality predictive model serves as a proof-of-concept for any type of AM machines to produce reliable parts with fewer quality hiccups while limiting the waste of both time and materials.

*Index Terms*—Additive manufacturing, Convolutional neural network, Quality monitoring system.

## I. INTRODUCTION

In smart manufacturing system (also known as Industry 4.0) [1], machines and robots must provide a high automation level with the ability to process information, enhance the yield of production [2, 3], visualize the performance in real-time [4], enable intelligent predictive maintenance system [5], and match service providers with customer demands [6]. Additive Manufacturing (AM) technology is a crucial component of the smart manufacturing system to enable flexible configuration and dynamic changing processes [7] to quickly adapt the products to new demands and potentially disrupt traditional supply chains. Additive manufacturing is applied in various sectors ranging from the fabrication of physical manufacturing prototypes [8, 9] to health care and biological products [10, 11]. Different printing methods have been developed to build 3D structures for personal or commercial purposes, including fused deposition modeling (FDM), stereolithography (SLA), digital light processing (DLP), and selective laser sintering (SLS). Despite the tremendous potentials for the AM methods to make custom-designed parts on-demand and with minimal material wastes [12], the widespread adoption of AM is hampered by poor process reliability and throughput due to lack of the condition-awareness of the AM process and automation. The parts built using current state-of-the-art AM machines have noticeable inconsistency in part and unpredictable mechanical properties [13, 14].

Currently, almost all AM machines have only limited sensing capabilities that are mostly inaccessible to the users or operating without any feedback measurement systems for correction during the process. Li et al. [15] used a sensor-based predictive model to predict the surface integrity of additively manufactured parts. Kousiatza and Karalekas [16] developed a monitoring system that predicts strain and temperature profiles using process condition data generated by thermocouples and optical sensors. However, the sensor-based monitoring systems need multiple sensors for diagnosing a single defect, while very few sensors can precisely monitor and recognize the product quality during the AM process. Ahn et al. [17] developed an analytical expression for surface roughness prediction using the geometrical information to investigate the effects of the static machine setting parameters on part quality. However, the predictive model has not considered the layer-by-layer nature of the AM process and the in-process variation during the printing process. Holzmond et al. [18] presented a non-destructive in-situ monitoring technique to detect the defects caused by residual stress. Greeff et al. [19] used an optical camera to monitor the filament fate rate and filament shredding

This paragraph of the first footnote will contain the date on which you submitted your paper for review. This work was supported in part by the National Science Foundation, the Louisiana Board of Regents, and Louisiana Consortium for Innovation in Manufacturing and Materials for funding this work through cooperative grant agreement OIA-1541079.

Y. Banadaki and N. Razaviarab are with the Department of Computer Science, Southern University and A&M College, Baton Rouge, Louisiana 70813, U.S.A. (e-mail: yaser_banadaki@subr.edu; nariman_razaviarab00 @subr.edu).

H. Fekrmandi is with the Department of Mechanical Engineering, South Dakota School of Mines and Technology, Rapid City, SD 57701, U.S.A. (e-mail: Hadi.Fekrmandi@sdsmt.edu).

S. Sharifi is with School of Electrical Engineering & Computer Science, Louisiana State University, Baton Rouge, LA 70803, U.S.A. (e-mail: sshari2@lsu.edu).

to study the required force. Plessis et al. [20] developed a monitoring system using X-ray computer tomography to study the effect of the nozzle blockage on large pores formation in build parts. Anderegg et al. [21] introduced an in-situ monitoring system to control the flow temperature and pressure of Fused Filament Fabrication (FFF) nozzle. Heras et al. [22] developed a system to detect whether the filament was moving forward properly to adjust the extrusion speed with the real speed of the AM process.

Future AM machines must be a smart system that can perform self-monitoring, self-calibrating, and quality self-controlling in real-time. The gap between the smart factory and existing manufacturing systems needs to be bridged concerning the automation, flexibility, and reconfigurability of AM machines in a computer-integrated manufacturing system. Machine learning [23, 24] can play an essential role in creating a multi-level of predictive models for the AM process. Several ML models have been explored to detect the defects in the AM process for specific processes and applications. Zhang et al. [3] focused on controlling powder quality in metal AM processes using an ML model and computational data obtained from the Discrete Element Method. Decost et al. [25] used ML methods to characterize, compare, and analyze powder feedstock materials and micrographs for the metal AM process. Stoyanov et al. [26] used an ML model to control the quality of 3D inkjet printing for designing electronic circuits. Inappropriate parameter settings could lead to building defects in the AM process, such as large pores and rough surfaces. Chen et al. [27] developed an ML model to optimize parameters of a Binder Jetting (BJ) process. Rao et al. [28] used Bayesian nonparametric analysis to identify failure modes and detect the onset of process anomalies in AM processes and found an optimal region of machine setting combination at high temperature, low layer thickness as well as high feed/flow rate ratio. Finally, Li et al. [29] used a support vector machine algorithm to identify the normal filament jam states in the FFF process. The studies on the in-situ monitoring and diagnosing of the AM process mostly used conventional data-driven methods such as support vector machine (SVM) [29-31] and hidden semi-Markov model (HSMM) [32], clustering method [33], and so on.

In this paper, we propose a deep learning-based computer vision system to monitor the quality characteristics of the additive manufacturing process. The residual pressure of the melted filament within the extrusion chamber may cause the material to overfill and underfill, which may lead to visible surface defects or invisible internal defects and, consequently, degradation in the quality and the mechanical performance of the printed parts. The monitoring system can predict and flag system failures before they happen to potentially manage the AM process, leading to a better chance of getting to the 100 percent yield. To implement the monitoring system, we first collected printing data for offline training of the predictive model and then evaluated and tested the predictive model for online monitoring of the AM process. The signal can be used as feedback to the machine to decide on "go" or "no go" based on the quality of the ongoing printing process and possibly suggest remedial actions by adjusting the process parameters. This ability turns 3D printers into essentially their own inspectors that keep track of printing interlayers to possibly eliminate the need to inspect parts after they are entirely built, thereby adds another layer of quality control [5, 6]. The deep learning-based computer vision system provides an automated, fast, consistent and more precise measure of printing quality to optimize the AM process, produce better parts with fewer quality hiccups, and limit the waste of time and materials. The proposed predictive model for the plastic AM process presented in this paper can serve as a proof-of-concept for other types of AM machines such as 3D bio-printers or polymer and liquid-based printers.

The rest of the paper is organized as follows. Section 2 presents the experimental arrangement, data collection, and training process to develop an efficient predictive model of automated fault detection and quality classification for the AM process. Section 3 discusses the performance measures of detecting and classifying the failure in AM process, including the model prediction of AM quality for different printing speeds and extruder temperatures as two controllable parameters with the greatest impact on the quality of built parts. The last section draws summarizing conclusions and future work.

## II. DEVELOPING A DEEP CONVOLUTIONAL NEURAL NETWORK MODEL

Figure 1 shows the data-driven predictive modeling procedure to implement the automated fault detection and quality classification system for the AM process by training a deep convolutional neural network.

### A. Data Collection from the AM process

An image acquisition system is established to capture the image of each layer of the part during the printing process that is used to prepare the dataset for training a non-contact quality predictive model based on machine vision technology. In AM technology, the geometry of parts forms layer by layer by joining of materials, and the geometrical deviation of each layer could affect the whole part quality. The machine vision system can capture frames for two main categories of surface and internal defects resulted from overfill and underfill of material, which may be visible in the part surface as excess material or a void, respectively [34]. The residual pressure of the melted filament within the extrusion chamber may cause the material to continue to be extruded, leading to excess material deposition and thus overfilling. To avoid these kinds of errors, the process parameters need to be optimized and dynamically adjusted the point at which extrusion should be stopped based on a feedback signal generated for the quality of the printing process in real-time. The possibility of adjusting parameters during the printing process can improve part accuracy and minimize defects because, for instance, a higher build chamber temperature may decrease part warping, but also may affect the surface finish [34]. Machine vision has recently emerged as a monitoring technology that can rapidly and automatically provide a huge number of samples for real-time control of product profiles in manufacturing processes.

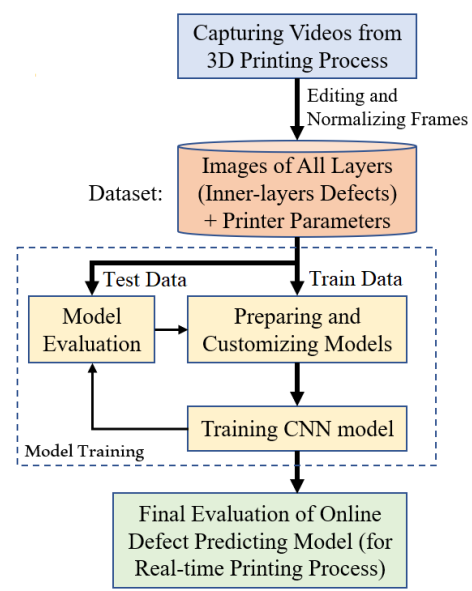

Fig. 1. Procedure to implement the fault detection and quality classification system for AM process, including collecting data, training the CNN model, and evaluating the model performance.

Figure 2(a) shows the experimental arrangement. The videos from the AM process are captured by a high-definition CCD camera (Lumens DC125). The training data is collected by filming the build of every layer in the AM process and converting the videos to frames. As such, the training data includes the occurrence of voids (internal bubbles) within the part that affects the structural integrity of the part and cannot be easily eliminated by post-processing [35]. The specimens are fabricated on a commercial desktop 3D printer (Creality3D Ender-3) to comply with the modern trends in the development of additive technologies or personal, desktop applications, which are within the affordable price range and are often able to produce parts to find applications in various walks of life. The printer works based on a fused deposition modeling method, in which a thermoplastic filament is heated to a semi-liquid state and deposited on a heated bed layer-by-layer to construct a 3D object [36]. The printer uses 1.75 mm-thick Polylactic Acid (PLA) build material to makes a solid to liquid transition by melting at extrusion temperatures from 185°C. The dataset is generated by printing the simple objects with different printing speeds (in mm/s) and different extruder temperatures (in °C) as machine settings of interest following the analysis in [37]. The printer allows us to change the printing speed from 50 mm/s to 1000 mm/s and the printing temperature from 185 C to 260 C. As such, to collect the training and test datasets, we print the objects by varying the printer parameters in 6 different printing speeds of 50 mm/s, 100 mm/s, 200 mm/s, 400 mm/s, 800 mm/s, 1000 mm/s, and four different printing temperatures of 185 C, 200 C, 230 C, 260 C.

Figure 2(b) shows sample frames of printing parts captured for training the CNN model. It can be observed that altering the speed and temperature of the AM process can significantly affect the quality of the printed parts. Increasing the speed and decreasing the temperature of the AM process lowers the quality of the printed part. The printer fails to print objects at the temperature of 185 C for the speeds above 200 mm/s because the extrusion temperature needs to be higher to melt the plastic quickly faster when the plastic is being pulled through the extruder. As such, the training dataset includes the videos of the printed parts with six different speeds and four temperatures that classifies into 21 categories. The videos of the AM process are converted to frames, and the rates of the frame extraction are adjusted for different printing speeds to ensure capturing the AM process of at least ten layers to includes images from the initial production of internal defects in the layer-by-layer deposition during the AM process. In other words, the training database includes the hidden information of defects as excess material or voids that are mostly invisible for human inspectors and can be very important for the mechanical performance of the printed parts. Then, the images are manually inspected to delete the images that the printer nozzle blocks the proper view of the printing area. After normalizing the intensity of images, 5000 images are chosen for the training process, and 100 images are randomly selected as test data for each class to evaluate the predictive model. For precise inspection for printing defects, the sample images with the size of 600×600 pixels are captured to be large enough to express small-sized defects in the AM process.

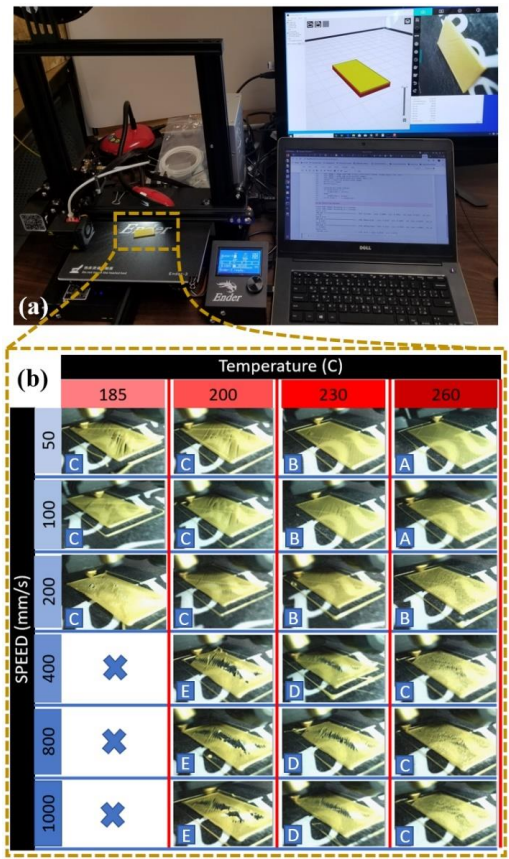

Fig. 2. Data collection and annotation for various printing parameters. (a) Experimental Setup, including 3D printer and AM monitoring system; and (b) Sample frames of printing parts captured for training the CNN model. Collecting data with six different flow rates of printing material and four different temperatures of printing material. The collected data annotated in five classes based on the printing quality, A: highest quality, and E: lowest quality. The crosses show the printer settings that completely fail to print any objects. Note: the figures show the quality of the finished surface of the printed specimens while the training database includes the hidden defect information in the inter-layers of the objects that is invisible for human inspector and can be very important for the mechanical performance of the printed parts.

*B. Training CNN model*

The conventional machine-learning techniques are not fully automated so that it needs to learn effective features and to extract feature vectors from input patterns through a feature extraction algorithm. This procedure requires human intervention in a training procedure that may affect the accuracy of the classification algorithm. In this paper, we used CNN (convolutional neural networks) architectures within a deep learning framework [38] that solve the shortcomings of the existing machine learning approaches. As a kind of machine learning and a particular type of neural network with deep layer architecture, CNN performs multilayer convolution to extract features and combine the features automatically at the same time on a single network. CNN extracts spatial features from low-level layers that are then passed to aggregation layers (convolutional, pooling, etc.) and additional layers of filters for extracting higher-order features (patterns) that are combined at the top layers, and fully connected (FC) layers in the output part of the network perform image interpretation and classification, as shown in Fig.3(a). As feature extraction and classification are simultaneously performed in a neural network, features fit for the classification are automatically carried out that further improving performance.

The image patches of the AM process are fed into a deep CNN for efficient AM quality detection, including three convolutional layers, three pooling layers, two dropout layers, and two fully connected layers. The composition of hidden layers is optimized to reach the optimal overall classification performance and training time. The composition of hidden layers relates to the number of convolution and pooling layers, the number of nodes in a convolution layer, and the kernel size of the pooling and convolution mask. The image size and the data size determine the optimal number of layers in DCNN such that the size of images and the number of classes affect the mask of layers and the number of nodes, respectively. The performance and reliability of CNN are directly associated with the number of sample data and the depth of layers. Without a public dataset, it is difficult to find AM images suitable for different scenarios of defect creation in the AM process, and thus increasing the depth of the network for a limited number of sample images results in over-fitting, further lowers the reliability of the model. The increase in the size of images allows expanding the depth of the neural network by adding more layers, possibly improving the CNN performance. However, more layers lead to an exponential increase in the computing cost, making the repetitive convolution-pooling structure necessary to be effectively parallelized to reduce the computing time.

A ReLU non-linear activation function is used for the input layer and hidden layers, while a logistic regression (softmax) function is implemented to generates a normalized exponential distribution for the final layer to obtain the final learning probability and predicted labels. A deep CNN has many hidden layers. To learn all the weights in the layers, the loss function is minimized by batch gradient descent algorithm that generally used to train a neural network to propagate an error by the chain rule. During the training steps, CNN learns optimal weights of all layers using forward- and backward- propagations through the neural network architecture. The architecture is employed by retraining a pre-trained model, the Inception-v3 [39], in the TensorFlow platform [40] that introduced as deep learning open-source software by Google to identify and classify images. TensorFlow has the advantages of high availability, high flexibility, and high efficiency. Transfer learning extracts existing knowledge learned from one environment to solve the other new problems such that the pre-trained CNNs take advantage of training with a lower amount of data for the new problem and significantly shortened the training time.

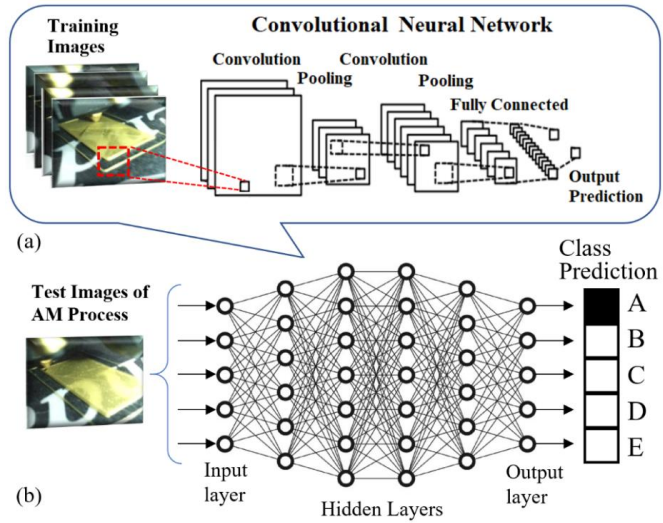

Fig. 3. Training and testing procedure of deep neural networks. (a) Training convolutional neural networks that includes procedures to extract spatial features to pass to aggregation layers (averaging and pooling), followed by higher-order features extraction that is combined at the top layer for AM fault detection and quality classification; and (b) Schematics of deep neural network including an input layer, multiple hidden layers, and an output layer that classifies test images of AM process to five classes of AM quality in real-time.

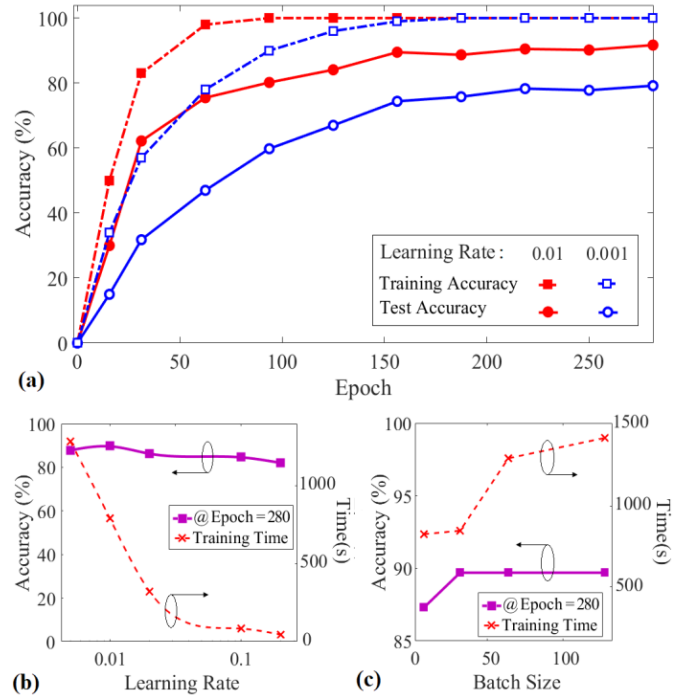

Fig. 4. Convergence of deep learning models for different number of epochs, learning rates, and batch size. (a) Training and test accuracies versus epoch for the DCNN predictive model of printing quality in AM process at a batch size of 8 and learning rate of 0.01; (b) Accuracy and training time of the DCNN model versus learning rate at epoch = 280 and batch size = 32; and (c) Accuracy and training time of the DCNN model versus learning rate at epoch = 280 and learning rate = 0.01.

To test and optimize the performance of the DCNN model, we conduct systematic convergence studies concerning the epochs, learning rate, and batch size. The train and test accuracies of the predictive model versus epoch for two different learning rates is shown in Fig. 4(a). We observe that both the train and the test accuracies increase by increasing the number of epochs, and the higher learning rate accelerates the convergence of the DCNN model. Another significant observation in Fig 4(a) is that the fluctuations of the test accuracies are very small as the number of iterations increases after 150 epochs, showing that the size of the datasets and the DCNN model are correctly selected, and the model is not suffering from overfitting.

The learning rate is the most critical hyperparameter for the neural network that affects how quickly our predictive model of the AM process can converge to the best accuracy. Figure 4(b) shows the plot of the learning rate against the model accuracy and training time. As the learning rate increase, the accuracy stops increasing and starts to decrease after 0.01. The maximum accuracy of 91% can be achieved at this learning rate for the batch size of 32. While choosing higher learning rates will raise the accuracy faster, it makes the optimization process unable to settle in the global minimum of the loss function, lowering the model accuracy. The training time is recorded for the experiments on the Intel® i5 desktop without parallel processing devices such as GPU acceleration.

Batch size is another important hyperparameters to tune in modern deep learning systems. Choosing a small batch size allows the model to start learning before having to see all the data, but it may not converge to the global optima, resulting in a smaller accuracy of the predictive model. As shown in Fig. 4(c), decreasing the batch size to 8 decreases the accuracy of our model to 87% as the model starts to bounce around the global optima. At a batch size of 32, the model accuracy raises to 91% with roughly the same computational training time for training the model. Increasing the batch size cannot lead to further improvement in the model accuracy or computational speedups in our non-parallel computer systems, and in many cases, depending on the size of training databases, it decreases the model generalization resulting in smaller model accuracy.

We calculate five metrics for the final evaluation of the predicting performance of the printing quality in AM process including $Precision = TP/(TP+FP)$, $Sensitivity = TP/(TP+FN)$, $Specificity = TN/(FP+TN)$, $F\text{-}score = 2 \times TP/(2 \times TP + FP + FN)$, and $Accuracy = (TP+TN)/(TP+FP+FN+TN)$, where $TP$, $TN$, $FP$, and $FN$ are, respectively, the true positive, true negative, false positive and false negative number of the printing objects being classified for each class. The precision can be viewed as a measure of a classifier's exactness and the sensitivity (or recall) as a measure of a classifier's completeness such that low precision and sensitivity indicate, respectively, many false positives and many false negatives. The specificity measures the proportion of correctly identified negatives, and the F1-score considers both precision and recall indicating the worse accuracy when it reaches 0, best corresponds to 1.

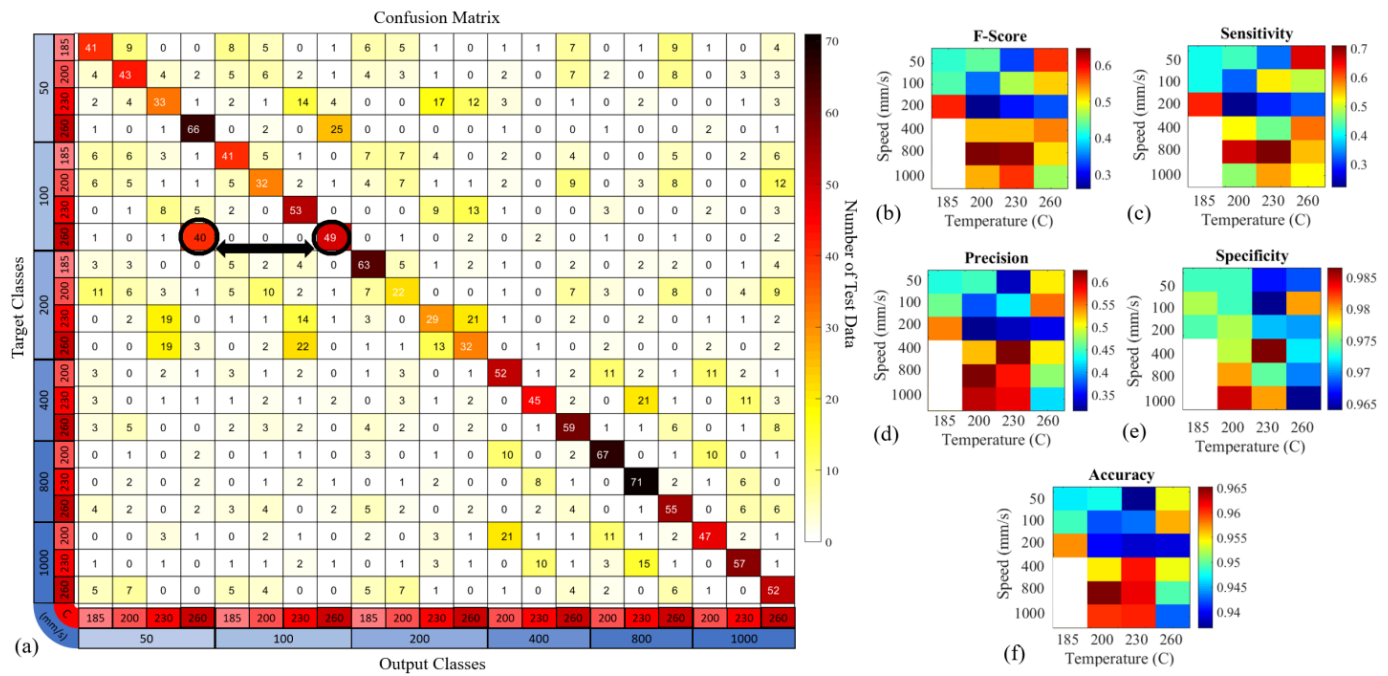

Fig. 5. Performance analysis for identifying 21 categories annotated based on temperatures and speeds settings of AM process. (a) Confusion matrix showing the exact number of correctly classified AM images and misclassified AM images; 5 statistic metrics for the model prediction to assign the printing part to 21 classes including (b) F-score, (c) sensitivity, (d) precision, (e) specificity, and (f) accuracy analysis.

## III. EVALUATING THE MODEL FOR AUTOMATED DETECTION OF THE AM QUALITY

### A. Performance Measures for 21 Classes of Speeds and Temperatures

Figure 5 shows the confusion matrix for the printed objects with six different speeds, and four temperatures classify into 21 classes [see Fig. 2(b)], which can guide humans to observe the dominant confusing classes for the classification model. For instance, the arrow in Fig. 5(a) shows that the classification algorithm has difficulty in correctly predicting the classes that are only distinguished based on the extrusion speeds, 50 mm/s and 100 mm/s, at the high temperature of 260 C. In the confusion matrix, the diagonal represents the correctly predicted number of each observation. Figure 5(b-f) depicts five measures that are computed for the performance analysis of the experiment. It can be noticed in Fig. 5(b) that the accuracies of all the 21 classes of AM process are above 93%, and the difference in accuracy among the classes is small as the accuracy refers to the true predictions (TP and TN) among the total validation. Besides the high accuracy or high specificity [Fig. 5(c)]; however, a good classifier must also demonstrate high performance for the other measures. Calculating the F-score, the sensitivity, and the precision of the predictive model [Fig. 5(d-f)] reveals that these classification factors can be as low as ~0.3, the dark blue regions, for the speed slower than 200 mm/s. Similarly, the maximum values of these three measures are not larger than 0.71, demonstrating the model prediction of many false positives and many false negatives. F1-scores of the classes are smaller than 0.5 for 11 out of 21 classes indicating low accuracy of the DCNN model for the classification of 21 AM process categories, especially for the printed classes with speed slower than 200 mm/s.

### B. Performance Measures for Predicting the Quality of AM Process

Extrusion speed and extrusion temperature are two controllable factors in the AM process that have a dominant impact on printing quality. Similarly, the predictive model of the AM process can be employed to detect the significant of the defects in the printing process and automatically grade the quality of the printing process. The signal can be used as feedback to the machine to decide whether the quality of printing is satisfactory for a given application and possibly suggest remedial actions by adjusting the process parameters. Figure 6 shows the performance analysis for the DCNN model to classify five quality grades (A to E) of the AM process annotated in Fig. 2(b). In the confusion matrix [Fig. 6(a)], out of 100 test images in each class, the maximum number of the correct prediction is 91 that belongs to the object printing with the quality grade of C, and the minimum number of the correct predictions is 81 for the object printing with the quality grade of E. Figure 6(b) depicts five measures that are computed for the performance analysis of the classification based on the quality grade of AM process. It can be noticed that the accuracies of the five quality classes of AM process, A to E, are 96%, 93.6%, 92%, 94.5%, and 94%, respectively. Similar to the classification of the 21 set-point classes, the specificity of the classifier is also high, equal to 98%, 96%, 92%, 97.5%, and 97.5% for grades A to E, respectively. However, the classification of the five quality classes of the AM process demonstrates better measures of the F-score, the sensitivity, and precision. As shown in Fig. 6(c), all the three measures are larger than 0.75, demonstrating the model prediction of a few false positives and a few false negatives. As such, the graph provides the comparison of two classification models, indicating significant improvement in the F-score, the sensitivity, and the precision of the classification of the five quality classes of AM process.

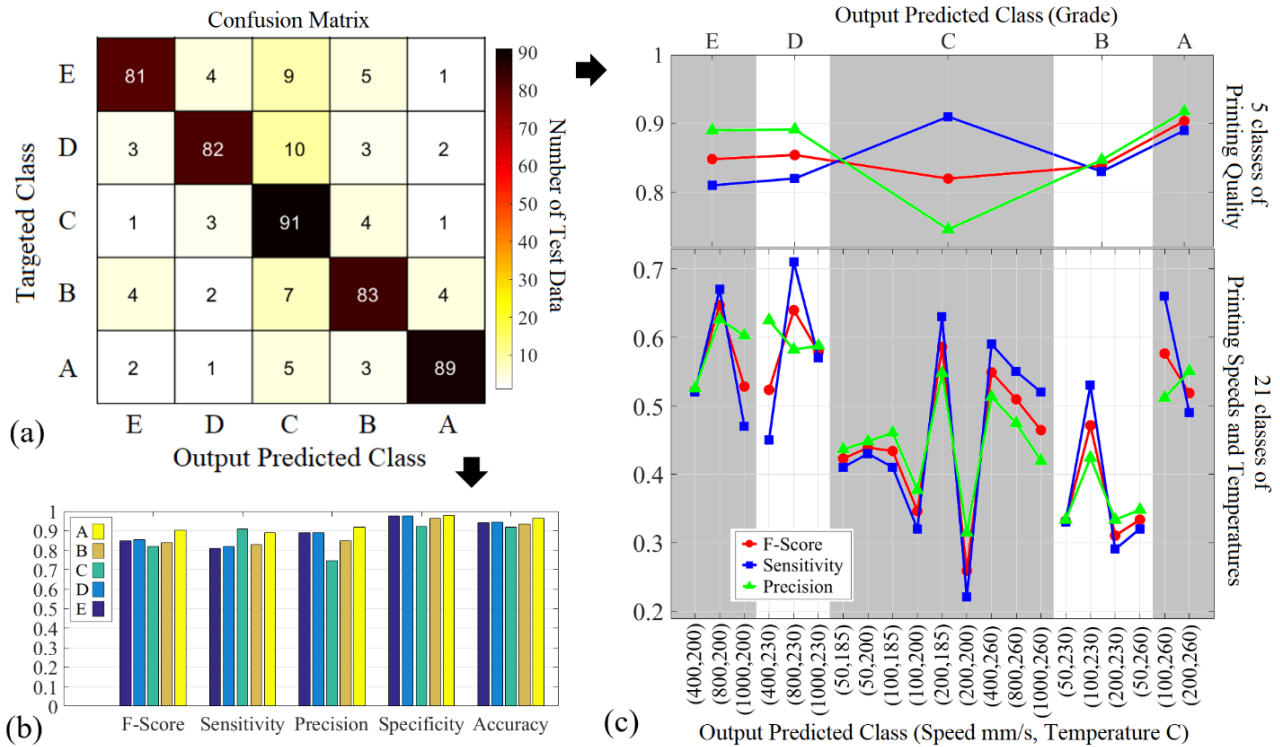

Fig. 6. Classification performance analysis for identifying five quality grades of the AM process. (a) Confusion matrix showing the detailed number of correctly classified and misclassified images of AM process; (b) 5 statistic metrics for the model prediction to classify the printing quality of AM process to five grades including F-score, sensitivity, precision, specificity, and accuracy analysis; and (c) Comparing F-score, sensitivity, and precision results for 21-class and 5-class classifications that shows the significant increase in these statistic metrics for A-E grades classification.

Figure 7 illustrates the model evaluation of the quality prediction of the AM process as a function of printing speed and temperature, including the true label of AM quality that corresponds to the annotation of the training data [Fig. 7(a)] and the predicted label of AM quality [Fig. 7(b)]. Comparing the two graphs indicates that the predicted model of AM quality can reach an accuracy larger than 90% in the dashed region, with an average accuracy of 98.2%, while outside of the dashed region has an average accuracy of 83%. It can be noticed that the DCNN model of AM process has a good prediction of the printing quality when printing with low speeds, ranging from 50 mm/s to 100 mm/s, regardless of the selected temperature of the AM process. It was further noticed that the model also has a good prediction of the printing quality when printing at a high temperature of 260 C for all printing speeds ranging from 50 mm/s to 1000 mm/s. In these regions, the model has high accuracy in classifying the printed parts into the printing quality grades.

## IV. SUMMARY, CONCLUSION, AND FUTURE WORKS

Additive manufacturing has tremendous potential to make a custom-designed part on-demand and with minimal material, but it is currently hampered by poor process reliability and throughput. In most industrial fields, AM defect inspection systems still depend on human inspection. Online process control using AM quality assurance increases the efficiency of factory automation as well as the possibility of a smart factory. Machine learning was announced as a critical component for the continuous growth of additive manufacturing technology. In this paper, we proposed a predictive model that can make a feedback signal for monitoring and controlling the AM process. We trained a deep convolutional neural network that results in the average accuracy of 94% and an average specificity of 96% in classifying the AM process into five quality classes. The three classifier measures of the F-score, the sensitivity, and precision are calculated all larger than 75%, demonstrating the model prediction with a few false positives and a few false negatives. The proposed predictive model for the plastic AM process presented in this paper serves as a proof-of-concept for other types of AM machines, such as 3D bio-printers, 3D polymer printers, and liquid-based printers. The automated inspection of AM quality can improve the speed, material waste, reliability, and productivity.

In this work, we used offline training with a fixed number of training samples for online predicting the defects and the quality of the AM process. Future work can be a smarter control system in the form of a closed-loop machine learning algorithm such that the model learns and adapts the important parameters as an AM machine is operating. The smarter 3D printer with the closed-loop ML model is trained itself by upgrading and improving its training samples over time to recognize any issues with the build to make proper adjustments along with corrections without operator intervention. This capability of the 3D printers ensures the high quality of AM parts by producing more reliable printed parts with fewer quality hiccups, shorter printing time, and less waste of materials. The suggested adaptive and self-improve 3D printer can observe new and different scenarios. The printer takes in new part build data, learns from experience, becomes smarter and more reliable, continuously improve the manufacturing process, automatically self-correct/compensate the deficiencies and thereby plays a vital role for future smart industries.

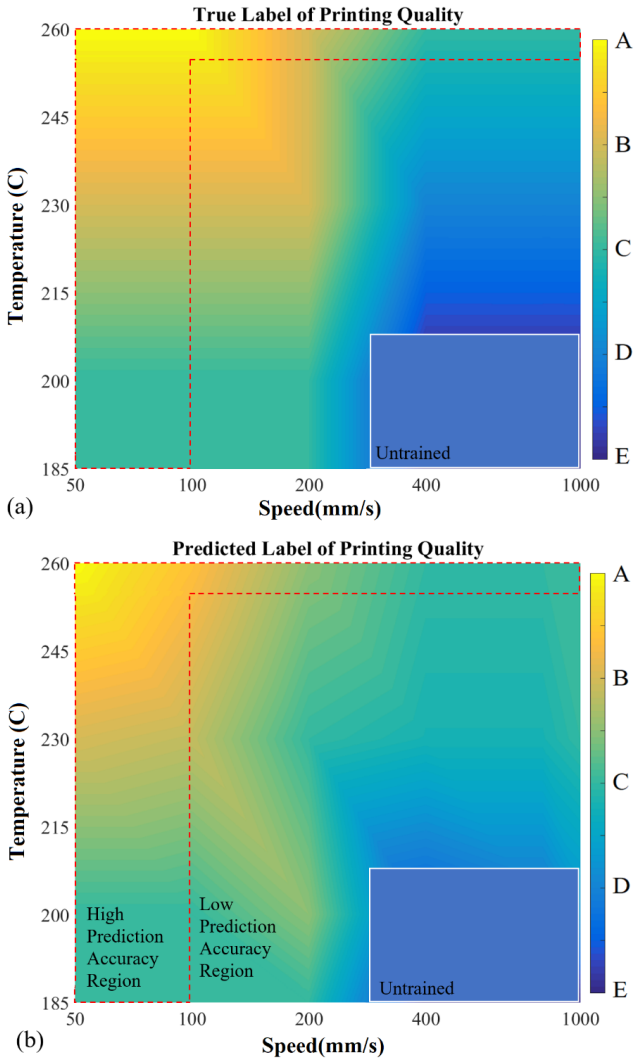

Fig. 7. Evaluation of predicted printing quality as a function of speed and temperature. (a) True label of AM quality versus speed and temperature, following the data annotation in Fig. 2(b); and (b) Predicted label of AM quality versus speed and temperature that includes the dashed region for the high prediction accuracy when it is compared with (a).

ACKNOWLEDGMENT

The authors acknowledge the National Science Foundation, the Louisiana Board of Regents, and Louisiana Consortium for Innovation in Manufacturing and Materials for funding this work through cooperative grant agreement OIA-1541079.

REFERENCES

[1] M. Hermann, T. Pentek, and B. Otto, "Design principles for industrie 4.0 scenarios," in *System Sciences (HICSS), 2016 49th Hawaii International Conference on*, 2016, pp. 3928-3937: IEEE.

[2] Y.-C. Lin *et al.*, "Development of Advanced Manufacturing Cloud of Things (AMCoT)—A Smart Manufacturing Platform," *IEEE Robotics and Automation Letters*, vol. 2, no. 3, pp. 1809-1816, 2017.

[3] W. Zhang, A. Mehta, P. S. Desai, and C. Higgs, "Machine learning enabled powder spreading process map for metal additive manufacturing (AM)," in *Int. Solid Free Form Fabr. Symp. Austin, TX*, 2017, pp. 1235-1249.

[4] P. Xu, H. Mei, L. Ren, and W. Chen, "ViDX: Visual diagnostics of assembly line performance in smart factories," *IEEE transactions on visualization and computer graphics*, vol. 23, no. 1, pp. 291-300, 2017.

[5] K.-S. Wang, Z. Li, J. Braaten, and Q. Yu, "Interpretation and compensation of backlash error data in machine centers for intelligent predictive maintenance using ANNs," *Advances in Manufacturing*, vol. 3, no. 2, pp. 97-104, 2015.

[6] X. Xue, Y.-M. Kou, S.-F. Wang, and Z.-Z. Liu, "Computational experiment research on the equalization-oriented service strategy in collaborative manufacturing," *IEEE Transactions on Services Computing*, vol. 11, no. 2, pp. 369-383, 2018.

[7] B. Scholz-Reiter, D. Weimer, and H. Thamer, "Automated surface inspection of cold-formed micro-parts," *Cirp annals-manufacturing technology*, vol. 61, no. 1, pp. 531-534, 2012.

[8] B. Bhushan and M. Caspers, "An overview of additive manufacturing (3D printing) for microfabrication," *Microsystem Technologies*, vol. 23, no. 4, pp. 1117-1124, 2017.

[9] C. K. Chua and K. F. Leong, *3D Printing and Additive Manufacturing: Principles and Applications (with Companion Media Pack) of Rapid Prototyping Fourth Edition*. World Scientific Publishing Company, 2014.

[10] C. L. Ventola, "Medical applications for 3D printing: current and projected uses," *Pharmacy and Therapeutics*, vol. 39, no. 10, p. 704, 2014.

[11] C. M. B. Ho, S. H. Ng, K. H. H. Li, and Y.-J. Yoon, "3D printed microfluidics for biological applications," *Lab on a Chip*, vol. 15, no. 18, pp. 3627-3637, 2015.

[12] S. Ford and M. Despeisse, "Additive manufacturing and sustainability: an exploratory study of the advantages and challenges," *Journal of Cleaner Production*, vol. 137, pp. 1573-1587, 2016.

[13] S. Guessasma, W. Zhang, J. Zhu, S. Belhabib, and H. Nouri, "Challenges of additive manufacturing technologies from an optimisation perspective," *International Journal for Simulation and Multidisciplinary Design Optimization*, vol. 6, p. A9, 2015.

[14] J.-Y. Dantan *et al.*, "Geometrical variations management for additive manufactured product," *CIRP Annals*, vol. 66, no. 1, pp. 161-164, 2017.

[15] Z. Li, Z. Zhang, J. Shi, and D. Wu, "Prediction of surface roughness in extrusion-based additive manufacturing with machine learning," *Robotics and Computer-Integrated Manufacturing*, vol. 57, pp. 488-495, 2019.

[16] C. Kousiatza and D. Karalekas, "In-situ monitoring of strain and temperature distributions during fused deposition modeling process," *Materials & Design*, vol. 97, pp. 400-406, 2016.

[17] D. Ahn, J.-H. Kweon, S. Kwon, J. Song, and S. Lee, "Representation of surface roughness in fused deposition modeling," *Journal of Materials Processing Technology*, vol. 209, no. 15-16, pp. 5593-5600, 2009.

[18] O. Holzmond and X. J. A. M. Li, "In situ real time defect detection of 3D printed parts," vol. 17, pp. 135-142, 2017.

[19] G. P. Greeff and M. J. A. M. Schilling, "Closed loop control of slippage during filament transport in molten material extrusion," vol. 14, pp. 31-38, 2017.

[20] A. du Plessis, S. G. le Roux, F. J. D. P. Steyn, and A. Manufacturing, "Quality Investigation of 3D printer filament using laboratory X-ray tomography," vol. 3, no. 4, pp. 262-267, 2016.

[21] D. A. Anderegg *et al.*, "In-situ monitoring of polymer flow temperature and pressure in extrusion based additive manufacturing," vol. 26, pp. 76-83, 2019.

[22] E. Soriano Heras, F. Blaya Haro, J. M. De Agustín del Burgo, M. Islán Marcos, and R. J. S. D'Amato, "Filament advance detection sensor for fused deposition modelling 3D printers," vol. 18, no. 5, p. 1495, 2018.

[23] N. M. Nasrabadi, "Pattern recognition and machine learning," *Journal of electronic imaging*, vol. 16, no. 4, p. 049901, 2007.

[24] R. S. Michalski, J. G. Carbonell, and T. M. Mitchell, *Machine learning: An artificial intelligence approach*. Springer Science & Business Media, 2013.

[25] B. L. DeCost, H. Jain, A. D. Rollett, and E. A. Holm, "Computer vision and machine learning for autonomous characterization of am powder feedstocks," *JOM*, vol. 69, no. 3, pp. 456-465, 2017.

[26] S. Stoyanov and C. Bailey, "Machine learning for additive manufacturing of electronics," in *Electronics Technology (ISSE), 2017 40th International Spring Seminar on*, 2017, pp. 1-6: IEEE.

[27] H. Chen and Y. F. Zhao, "Learning Algorithm Based Modeling and Process Parameters Recommendation System for Binder Jetting Additive Manufacturing Process," in *ASME 2015 International Design Engineering Technical Conferences and Computers and Information in Engineering Conference*, 2015, pp. V01AT02A029-V01AT02A029: American Society of Mechanical Engineers.

[28] P. K. Rao, J. P. Liu, D. Roberson, Z. J. Kong, and C. Williams, "Online real-time quality monitoring in additive manufacturing processes using heterogeneous sensors," *Journal of Manufacturing Science and Engineering*, vol. 137, no. 6, p. 061007, 2015.

[29] Y. Li, W. Zhao, Q. Li, T. Wang, and G. J. S. Wang, "In-Situ Monitoring and Diagnosing for Fused Filament Fabrication Process Based on Vibration Sensors," vol. 19, no. 11, p. 2589, 2019.

[30] J. S. Kim, C. S. Lee, S.-M. Kim, S. W. J. I. J. o. P. E. Lee, and M.-G. Technology, "Development of data-driven in-situ monitoring and diagnosis system of fused deposition modeling (FDM) process based on support vector machine algorithm," vol. 5, no. 4, pp. 479-486, 2018.

[31] H. Wu, Y. Wang, and Z. J. T. I. J. o. A. M. T. Yu, "In situ monitoring of FDM machine condition via acoustic emission," vol. 84, no. 5-8, pp. 1483-1495, 2016.

[32] H. Wu, Z. Yu, and Y. J. T. I. J. o. A. M. T. Wang, "Real-time FDM machine condition monitoring and diagnosis based on acoustic emission and hidden semi-Markov model," vol. 90, no. 5-8, pp. 2027-2036, 2017.

[33] J. Liu, Y. Hu, B. Wu, and Y. J. J. o. M. P. Wang, "An improved fault diagnosis approach for FDM process with acoustic emission," vol. 35, pp. 570-579, 2018.

[34] A. Peng, X. J. A. i. I. S. Xiao, and S. Sciences, "Investigation on reasons inducing error and measures improving accuracy in fused deposition modeling," vol. 4, no. 5, 2012.

[35] M. K. Agarwala, V. R. Jamalabad, N. A. Langrana, A. Safari, P. J. Whalen, and S. C. J. R. p. j. Danforth, "Structural quality of parts processed by fused deposition," 1996.

[36] L. Chong, S. Ramakrishna, and S. Singh, "A review of digital manufacturing-based hybrid additive manufacturing processes," *The International Journal of Advanced Manufacturing Technology*, vol. 95, no. 5-8, pp. 2281-2300, 2018.

[37] A. H. Peng and Z. M. Wang, "Researches into influence of process parameters on FDM parts precision," in *Applied Mechanics and Materials*, 2010, vol. 34, pp. 338-343: Trans Tech Publ.

[38] S. Hoo-Chang *et al.*, "Deep convolutional neural networks for computer-aided detection: CNN architectures, dataset characteristics and transfer learning," *IEEE transactions on medical imaging*, vol. 35, no. 5, p. 1285, 2016.

[39] C. Szegedy, V. Vanhoucke, S. Ioffe, J. Shlens, and Z. Wojna, "Rethinking the inception architecture for computer vision," in *Proceedings of the IEEE conference on computer vision and pattern recognition*, 2016, pp. 2818-2826.

[40] M. Abadi *et al.*, "Tensorflow: a system for large-scale machine learning," in *OSDI*, 2016, vol. 16, pp. 265-283.